# Bottom-up Instance Segmentation using Deep Higher-Order CRFs

Anurag Arnab
anurag.arnab@eng.ox.ac.uk

Philip H.S. Torr
philip.torr@eng.ox.ac.uk

Department of Engineering Science
University of Oxford
United Kingdom

## Abstract

Traditional Scene Understanding problems such as Object Detection and Semantic Segmentation have made breakthroughs in recent years due to the adoption of deep learning. However, the former task is not able to localise objects at a pixel level, and the latter task has no notion of different instances of objects of the same class. We focus on the task of Instance Segmentation which recognises and localises objects down to a pixel level. Our model is based on a deep neural network trained for semantic segmentation. This network incorporates a Conditional Random Field with end-to-end trainable higher order potentials based on object detector outputs. This allows us to reason about instances from an initial, category-level semantic segmentation. Our simple method effectively leverages the great progress recently made in semantic segmentation and object detection. The accurate instance-level segmentations that our network produces is reflected by the considerable improvements obtained over previous work at high $AP^r$ IoU thresholds.

## 1 Introduction

Object detection and semantic segmentation have been two of the most popular Scene Understanding problems within the Computer Vision community. Great advances have been made in recent years due to the adoption of deep learning and Convolutional Neural Networks [19, 22, 32]. In this paper, we focus on the problem of *Instance Segmentation*. Instance Segmentation lies at the intersection of Object Detection – which localises different objects at a bounding box level, but does not segment them – and Semantic Segmentation – which determines the object-class label of each pixel in the image, but has no notion of different instances of the same class. As shown in Figure 1, the task of instance segmentation localises objects to a pixel level.

Many recent instance segmentation works have built on the "Simultaneous Detection and Segmentation" (SDS) approach of Hariharan *et al*. [13]. These methods [4, 5, 13, 14] generate object proposals [1], classify each proposal into an object category, and then refine the bounding box proposal into a segmentation of the primary object the proposal contains. However, because these methods localise an object before segmenting it, they are restricted by the quality of the initial proposal. Moreover, the proposal generator [1] these works used takes about 40 seconds to process an image.

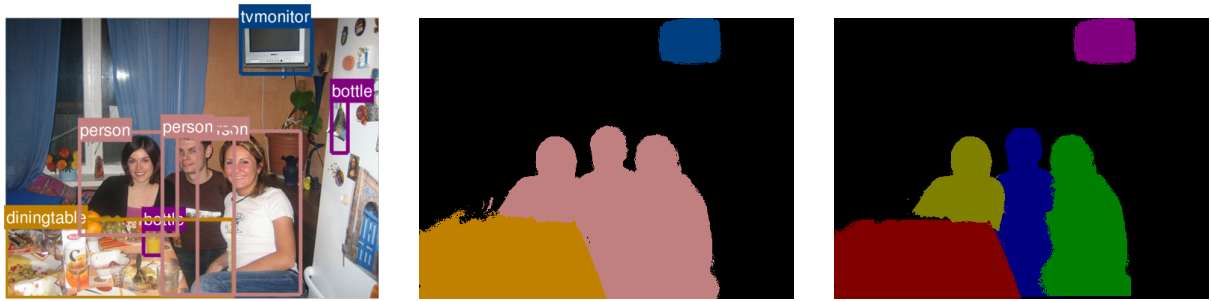

(a) Object Detection     (b) Semantic Segmentation     (c) Instance Segmentation

Figure 1: Our method uses the results of an object detector and the original image (a) to produce a category-level semantic segmentation (b). The semantic segmentation result, object detection bounding boxes, and recalibrated detection scores (§3.2) are then used to refine the semantic segmentation into an instance segmentation (c). The semantic segmentation, in which a large number of pixels have been correctly labelled as "person", does not tell us how many people there are and where each person is in the image. The instance segmentation answers these questions. Also note how our method is not affected by the false-positive "bottle" detections.

We present a different approach to instance segmentation, where we initially perform a category-level, semantic segmentation of the image before reasoning about the different instances in the scene. We can identify instances because our semantic segmentation network incorporates a Higher Order Conditional Random Field (CRF) which uses cues from the output of an object detector. This CRF is fully differentiable and can be inserted as a layer of a neural network [2, 37]. As a result, it can be used as a simple extension to an existing semantic segmentation network to perform the related task of instance segmentation.

Our simple, bottom-up method is able to effectively leverage the progress made by state-of-the-art semantic segmentation and object detection networks to perform the related task of instance segmentation. We show this by evaluating our approach on the PASCAL VOC 2012 [8] dataset, using the standard $AP^r$ measure [13]. The fact that our system is able to produce accurate instance segmentations is reflected by the considerable improvements that we make over previous work at high overlap thresholds.

## 2 Related Work

Instance segmentation is closely related to the widely-studied problems of semantic segmentation and object detection. We thus review these two tasks first.

**Semantic Segmentation:** This task was traditionally performed by extracting dense features from an image and classifying them on a per-pixel or per-region basis [16, 31]. The individual predictions made by these classifiers were often noisy as they lacked global context. As a result, they were post-processed with a CRF whose priors encourages nearby pixels, and pixels of similar appearance, to share the same class label. The CRF of [31] initially contained only unary terms (obtained from the classifier) and 8-connected pairwise terms. Subsequent improvements included introducing densely-connected pairwise potentials [18], using higher order potentials encouraging consistency over cliques larger than two pixels [17, 35], modelling the co-occurrence of labels [20, 35] and utilising the results of object detectors [21, 33, 36].

Using CNNs for semantic segmentation can greatly improve the unary, per-pixel predictions, as shown by [27]. Chen *et al*. showed further improvements by post-processing the CNN output with a CRF [3]. Current state-of-the-art methods [2, 25, 37] incorporate inference of a CRF as layers within a deep network which perform the differentiable mean field inference algorithm. In our work, we extend the end-to-end trainable, higher-order, object detection potentials proposed by [2] to the task of instance segmentation. Additionally, unlike the aforementioned works which always assume a fixed number of labels in the CRF at all times, our system includes a CRF where the number of labels varies per input image.

**Object Detection:** Part-based models [9] were popular before CNNs, which had been shown to excel in the task of image classification [19, 32], were adapted to object detection. Recent detection algorithms are based on using CNNs to classify object proposals and then performing bounding box regression to refine the initial proposals' bounding boxes [10, 11, 12, 28]. This pipeline can be accelerated by computing convolutional features over the whole image, and then pooling features for each proposal [11, 15]. Object proposals were initially hand-crafted [1, 34], but this step has now been integrated into CNNs [28].

**Instance Segmentation:** This task was popularised by Hariharan *et al*. [13]. Their approach was similar to object detection pipelines in that region proposals were first generated and classified into different object categories before using bounding box regression as post-processing. A class-specific segmentation was then performed to simultaneously detect and segment an object. However, approaches that segment instances by refining detections [4, 5, 13, 14] are inherently limited by the quality of the initial proposals, and the systems cannot recover from errors made by the initial detector. Furthermore, since these methods generate and classify multiple overlapping region proposals per image, they cannot actually produce segmentation maps of the form shown in Figures 1(b) and (c) [24]. Liang *et al*. [23] address some of these shortcomings by iteratively refining initial object proposals, whilst [7] generates box-proposals, creates masks from these proposals and then classifies these masks in a single network trained with three loss functions summed together.

A proposal-free method was recently presented by [24] where a semantic segmentation of the image is first performed using the network of [3]. Thereafter the category-level segmentation output, along with CNN features, are used to predict instance-level bounding boxes for the objects in the image. The number of instances of each object category are also predicted in order to facilitate the final clustering step. An alternate method was proposed by [29] where a Recurrent Neural Network outputs an object instance at each time step. This method, however, has not been evaluated on multiple classes.

Our method also first performs a bottom-up, semantic segmentation of the image. However, our semantic segmentation network uses the outputs of an object detector as an additional cue in its final CRF layer. During inference, each object hypothesis from the detector is evaluated in light of other energies in the CRF. Thus, the relative score of false positive detections can be decreased, and correct detections increased. These rescored detections are then used to identify instances from the initial semantic segmentation. The fact that our system combines information from semantic segmentation and object detection networks allows us to bypass the complexity of the "instance bounding box" and "instance number" prediction steps used by [24].

## 3 Proposed Approach

Our proposed method first performs a semantic segmentation of the input image, classifying each pixel into one of $K+1$ categories where $K$ is the number of foreground classes. The resulting semantic segmentation is then refined into an instance-level segmentation, where the object class of each instance segment is obtained from the previous semantic segmentation. Both of these stages, while conceptually different, are fully differentiable and the entire system can be implemented as a neural network.

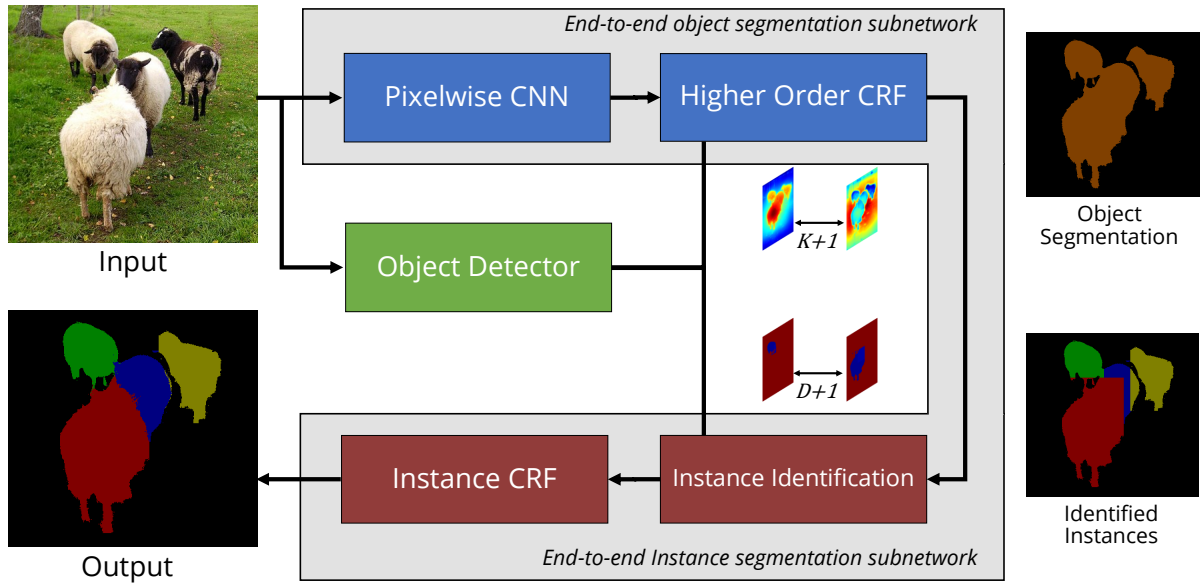

Figure 2: Overview of our end-to-end method. Our system first computes a category-level semantic segmentation of the image. A CRF with higher-order detection potentials is used to obtain this result in a semantic segmentation subnetwork. This results in a $W \times H \times (K+1)$ dimensional volume where $K$ is the number of foreground classes. $W$ and $H$ are the image's width and height respectively. The original detections and the detector confidences recalibrated by the Higher Order CRF are used to identify instances in the image, producing a $W \times H \times (D+1)$ dimensional volume where $D$ is the number of detections (variable per image). Thereafter, another Instance CRF is used to compute the final result.

Figure 2 shows our category-level segmentation network, consisting of a pixelwise CNN [27] followed by a fully differentiable CRF with object detection potentials, as described in Section 3.2. The object detection results that our CRF takes in, as well as the recalibrated detection scores that it outputs, allows us to identify object instances, and obtain a final instance-level segmentation in the next stage of our pipeline. This is described in Section 3.3. However, we first review Conditional Random Fields and introduce the notation used in this paper.

## 3.1 Conditional Random Fields

Assume an image $\mathbf{I}$ with $N$ pixels, indexed $1, 2 \ldots N$, and define a set of random variables, $X_1, X_2, \ldots, X_N$, one for every pixel. In a generic labelling problem, we assign every pixel a label from a predefined set of labels $\mathcal{L}$ such that each $X_i \in \mathcal{L}$. Let $\mathbf{X} = [X_1\ X_2\ \ldots X_N]^T$. In this case, any particular assignment $\mathbf{x}$ to $\mathbf{X}$ is a solution to our labelling problem.

In the case of semantic segmentation, we assign each pixel, $X_i$, a label corresponding to object categories such as "person" and "car". In instance segmentation, the labels are drawn from the product label space of object categories and instance numbers. Examples are "person_1" and "person_2" as shown by the different colours in Figure 1.

Given a graph $G$ where the vertices are from $\{\mathbf{X}\}$ and connections between these variables are defined by edges, the pair $(\mathbf{I}, \mathbf{X})$ is modelled as a CRF defined by $\Pr(\mathbf{X} = \mathbf{x}|\mathbf{I}) = (1/Z(\mathbf{I})) \exp(-E(\mathbf{x}|\mathbf{I}))$. The term, $E(\mathbf{x}|\mathbf{I})$ is the energy of the labelling $\mathbf{x}$ and $Z(\mathbf{I})$ is the

data-dependent partition function. From here onwards, we drop the conditioning on **I** to keep the notation uncluttered. The energy of an assignment is defined over the set of cliques $\mathcal{C}$ in the graph $G$ as

$$E(\mathbf{x}) = \sum_{c \in \mathcal{C}} \psi_c(\mathbf{x}_c), \tag{1}$$

where $\mathbf{x}_c$ is the vector formed by selecting elements of $\mathbf{x}$ that correspond to random variables belonging to the clique $c$, and $\psi_c(.)$ is the cost function for a clique $c$.

The energy function considered by many semantic segmentation works, $E(\mathbf{x}) = \sum_i \psi_i^U(x_i) + \sum_{i<j} \psi_{ij}^P(x_i, x_j)$, consists of only unary (cliques of size one) and densely-connected pairwise potentials (cliques of size two) [3, 18, 37]. When dense pairwise potentials are used in a CRF to obtain higher accuracy, exact inference is not tractable and approximate inference methods such as mean field are used [18].

## 3.2 Higher Order Detection Potentials

In addition to the standard unary and densely-connected pairwise terms, we also include a higher order term based on object detections. Intuitively, a state-of-the-art object detector [11, 28] can help segmentation performance in cases where the detector correctly predicts an object at a location in which the pixels have been misclassified by the unary classifier, and thus reduce the recognition error [6]. Arnab *et al.* [2] formulated object detection potentials in a manner that is amenable to the differentiable mean field inference algorithm so that it could be integrated into a neural network, and showed that it does indeed improve semantic segmentation performance. We review the detection potentials described in [2] as they also help us to reason about instances in an input image, a problem not considered by [2].

Assume that we have $D$ object detections for an input image (this number varies for every image), and that the $d^{th}$ detection is of the form $(l_d, s_d, F_d, B_d)$ where $l_d \in \mathcal{L}$ is the object class label of the detected object, $s_d$ is the detector's confidence score, $F_d \subset \{1, 2, \ldots N\}$ is the set of indices of the pixels belonging to the foreground of the detection and $B_d \subset \{1, 2, \ldots N\}$ is the set of indices falling in the detection's bounding box. The foreground is obtained from a foreground/background segmentation method such as GrabCut [30] and indicates a rough segmentation of the detected object. The detection potentials should encourage the set of pixels represented by $F_d$ to take the object class label, $l_d$. However, this should be a soft constraint, since the foreground segmentation could be inaccurate and the entire detection itself could be a false-positive. The detection $d$ should also be identified as invalid if other energies in the CRF strongly suggest that many pixels in $F_d$ do not take the label $l_d$.

This is formulated by introducing a latent binary random variable, $Y_1, Y_2 \ldots Y_D$ for every detection. If the $d^{th}$ detection has been found to be valid after inference, $Y_d$ will be set to 1, and 0 otherwise. The final value of $Y_d$ is determined probabilistically by mean field inference.

All latent $Y_d$ variables are added to the CRF which previously only contained $X_i$ variables. Let each $(\mathbf{X}_d, Y_d)$, where $\{\mathbf{X}_d\} = \{X_i \in \{\mathbf{X}\} | i \in F_d\}$, form a clique in the CRF. An assignment $(\mathbf{x}_d, y_d)$ to the clique $(\mathbf{X}_d, Y_d)$ has the energy:

$$\psi_d^{Det}(\mathbf{X}_d = \mathbf{x}_d, Y_d = y_d) = \begin{cases} w_l \frac{s_d}{|F_d|} \sum_{i=1}^{|F_d|} [x_d^{(i)} = l_d] & \text{if } y_d = 0, \\ w_l \frac{s_d}{|F_d|} \sum_{i=1}^{|F_d|} [x_d^{(i)} \neq l_d] & \text{if } y_d = 1, \end{cases} \tag{2}$$

where $x_d^{(i)}$ is the $i^{th}$ element in vector $\mathbf{x}_d$, $[.]$ is the Iverson bracket and $w_l$ is a class-specific, learnable weight parameter. This potential encourages consistency among $X_d^{(i)}$ variables and $Y_d$ since it encourages $X_d^{(i)}$ variables to take the label $l_d$ when $Y_d$ is 1, and also encourages $Y_d$ to be 0 when many $X_d^{(i)}$ variables do not take the label $l_d$.

A unary potential for the latent $Y$ variables is also included, which is initialised to the confidence score $s_d$ of the object detector (in addition to the joint potential with the $X$ variables). This initial confidence then changes throughout CRF inference, depending on the other potentials. The final probability of the latent $Y$ variable can thus be viewed as a recalibrated detection score, which gets adjusted during inference based on how much it agrees with the segmentation unaries and pairwise potentials. Note that this "recalibration property" of the latent $Y$ variables was not considered by [2]. The final form of the energy function (Eq 1), is thus

$$E(\mathbf{x}) = \sum_i \psi_i^U(x_i) + \sum_{i<j} \psi_{i,j}^P(x_i,x_j) + \sum_d \psi_d^{Det}(\mathbf{x}_d, y_d) + \sum_d \psi_d^U(y_d). \quad (3)$$

## 3.3 Instance Identification and Refinement

Once we have a category-level segmentation of the image, each pixel still needs to be assigned to an object instance. We assume that each object detection represents a possible instance. Since there are $D$ object detections (where $D$ varies for every image), and some pixels do not belong to any object instance, but are part of the background, we have a labelling problem involving $D+1$ labels. Our set of labels, $\mathcal{D}$, is thus $\{1, 2, \ldots D+1\}$.

A naïve solution to the problem of recovering instances from a category-level segmentation would be as follows: If a pixel falls within the bounding box of a detection, and its category label agrees with the class predicted by the object detector, then it is assigned to the instance represented by that detection. However, this method cannot deal with overlapping bounding boxes which are typical when objects occlude one another (as in Figs. 1 and 2).

Instead, if a pixel falls within the bounding box of a detection, we assign the pixel to that instance with a probability proportional to the rescored detection (obtained from the probability of the latent $Y$ variable after inference) and the semantic segmentation confidence for that class, as shown in Equation 4

$$\Pr(v_i = k) = \begin{cases} \frac{1}{Z(\mathbf{Y},\mathbf{Q})} Q_i(l_k) \Pr(Y_k = 1) & \text{if } i \in B_k \\ 0 & \text{otherwise.} \end{cases} \quad (4)$$

Here, $v_i$ is a multinomial random variable indicating the "identified instance" at pixel $i$ and takes on labels from $\mathcal{D}$, $Q_i(l)$ is the output of the initial category-level segmentation stage of our network and denotes the probability of pixel $i$ taking the label $l \in \mathcal{L}$, and $Z(\mathbf{Y}, \mathbf{Q})$ normalises the probability in case there are multiple bounding boxes overlapping the same pixel. For this formulation to be valid, we also add another detection, $d_0$, denoting background pixels that are not overlapped by any detection. This then acts as the unary potentials of another CRF with the energy:

$$E(\mathbf{v}) = \sum_i \psi_i^U(v_i) + \sum_{i<j} \psi_{i,j}^P(v_i, v_j) \qquad \psi_i^U(v_i) = -\ln \Pr(v_i). \quad (5)$$

This Instance CRF contains unary, $\psi_i^U(v_i)$, and densely-connected pairwise terms, $\psi_{i,j}^P(v_i, v_j)$, encouraging appearance and spatial consistency [18]. These priors are valid in the case of instance segmentation as well. We then perform mean field inference on this final Instance CRF to get our final output. Note that this final, pairwise CRF is dynamic – the $D+1$ labels that it deals with are different for every image, where $D$ is equal to the number of object detections obtained after non-maximal suppression. This differs with the CRFs considered in semantic segmentation literature which always have a fixed number of labels for all images in the dataset [17, 18, 35, 37]. Zheng *et al.* [37] showed how the iterative mean field inference algorithm can be unrolled into a series of differentiable operations and be implemented as a recurrent neural network. We employ a similar method so that our dynamic CRF is fully differentiable and can thus be incorporated into a neural network. This is achieved by using CRF weights that are not class-specific, and can thus generalise to any number of labels per image. A label here refers to an instance associated with a detection and has no semantic meaning. For example, in Fig. 1, label "1" was associated with a dining table, whilst in Fig. 2 it is a sheep. It therefore does not make sense to have class-specific weights in any case.

This approach to instance segmentation is bottom-up, in the sense that we first perform category-level segmentation of each pixel before predicting instances. However, the fact that we used object detection cues in performing our semantic segmentation helps us to subsequently reason about instances. Moreover, the fact that we used results from both semantic segmentation and object detection makes sense since instance segmentation can be regarded as being at the intersection of these two more common problems.

## 4 Experimental Results and Analysis

We first describe the dataset and our experimental setup (§4.1), before examining the efficacy of our detection potentials (§4.2) and comparing our approach with other recent instance segmentation methods (§4.3).

### 4.1 Experimental Setup

Following previous work, we evaluate our instance segmentation performance on the PASCAL VOC 2012 [8] validation set which comprises of 1449 images with high-quality annotations. There is no test server for instance segmentation to evaluate on the test set. We use the standard $AP^r$ measure for evaluation [13], which computes the mean average precision under different Intersection over Union (IoU) scores between the predicted and ground truth segmentations (rather than IoU between bounding boxes, as in object detection). In object detection, a predicted bounding box is considered correct if the IoU between it and the ground-truth bounding box is greater than 0.5. Here, we consider different overlap thresholds between the predicted and ground-truth segments, since higher overlap thresholds require the segmentations to be more precise and are therefore better indicators of segmentation performance. We also quote the $AP^r_{vol}$ measure, which is the mean $AP^r$ score for overlap thresholds ranging from 0.1 to 0.9 in increments of 0.1 [13]. The detailed annotations in the VOC 2012 dataset allows us to reliably evaluate the $AP^r$ of our method at high overlap thresholds.

We first trained a network for semantic segmentation, in a similar manner to [27] and [37]. Although the trained models from these two authors have been made publicly available, we could not use them since they had been trained on part of the VOC validation set. We first trained a fully convolutional network [27] (finetuned from VGG-16 trained on ImageNet [32]) using VOC 2012 training data, augmented with images from the Semantic Boundaries Dataset [13] which do not overlap with the VOC Validation set, and the Microsoft COCO dataset [26].

To this pretrained network, we added the Higher Order CRF, and finetuned with a learning rate of $10^{-11}$ using only VOC 2012 data which is finely annotated. The learning rate is low since the loss was not normalised by the number of pixels in the training image. When training our Higher Order CRF, we use the publicly available Faster-RCNN object detection framework [28]. The semantic segmentation performance on the VOC validation set is a mean IoU of 73.4% when using only pairwise potentials in the CRF. This rises to 75.3% when using detection potentials as well. The authors of [2] observed a similar increase when using detection potentials.

Table 1: Comparison of instance segmentation performance against baselines on the VOC 2012 Validation Set of 1449 images. We report the $AP^r$ measure at different IoU thresholds.

| Method | $AP^r$ at 0.5 | $AP^r$ at 0.6 | $AP^r$ at 0.7 | $AP^r_{vol}$ |
| --- | --- | --- | --- | --- |
| Baseline without detection potentials | 54.6 | 48.5 | 41.8 | 50.0 |
| Baseline with detection potentials, but $Y$ variables ignored | 57.5 | 51.6 | 44.5 | 52.4 |
| Full system with detection potentials, and recalibrated detection scores from $Y$ variables | 58.3 | 52.4 | 45.4 | 53.1 |

## 4.2 Effect of detection potentials

To analyse the effect of our detection potentials, we evaluate our system by disabling the higher order potentials in our Semantic Segmentation CRF. Because our detection potentials are disabled, we can no longer use the output of the latent $Y$ variables as our recalibrated detection score, and use the original detector's confidence score instead. As shown in Table 1, the result is that the $AP^r$ at 0.5 is 3.7% lower and the $AP^r_{vol}$ is 3.1% lower.

The detection potentials improve instance segmentation performance as they improve both our initial semantic segmentation, and also recalibrate the detector's initial scores. To decouple these two effects, we then include detection potentials in our Segmentation CRF, but ignore the latent $Y$ variables output by the CRF and use the detector's original confidence score instead as the input to the instance segmentation network. The second row of Table 1 shows that this second baseline has an $AP^r$ at 0.5 and an $AP^r_{vol}$ that are respectively 0.8% and 0.7% lower than our final method. The difference between the final method and this second baseline is due to the score recalibration performed by the detection potentials. The two baselines differ in performance since the second baseline uses a semantic segmentation (mean IoU of 75.3%) which is better than the other (73.4%) due to its use of detection potentials. The large gap in their overall instance segmentation performance emphasises the importance of a good initial semantic segmentation for our method, and underlines the benefits of a bottom-up approach to instance segmentation.

From the difference in performance between the baseline that does not fully make use of detection potentials (Row 1 of Table 1), and our final method (Row 3 of Table 1), we can conclude that our higher order detection potentials are an integral part of our system.

## 4.3 Comparison to other current methods

In Table 2, we compare our methods to other approaches that have been evaluated on the same dataset and reported $AP^r$ results at multiple IoU thresholds. The SDS method of [13] was evaluated on the VOC Validation set by [4]. Our approach significantly outperforms both [13] and [4] which initially perform object detection, and refine these detections into an instance segmentation. In fact, even our baselines in Table 1 outperform them. The model of [24] differs from the other two works in that the first stage of their pipeline is to also perform a category-level segmentation of the image. However, their method does not make use of any information from object detectors, and our approach performs better, particularly at high IoU thresholds where the difference in $AP^r$ at an IoU threshold of 0.9 is 4.4%. This indicates that our instance segmentations are more precise. The fact that our $AP^r_{vol}$ is only 0.8% greater, despite our $AP^r$ being significantly larger at high thresholds suggests that the method of [24] has a higher $AP^r$ at lower thresholds (below 0.5, which were not reported by [24]). This indicates that their method identifies more instances than ours, but the fact that their $AP^r$ at higher thresholds is lower than ours also suggests that their segmentations are not as accurate.

Table 2: Comparison of instance segmentation performance to other methods on the VOC 2012 Validation Set. We report the $AP^r$ measure at five different IoU thresholds.

| Method | $AP^r$ at 0.5 | $AP^r$ at 0.6 | $AP^r$ at 0.7 | $AP^r$ at 0.8 | $AP^r$ at 0.9 | $AP^r_{vol}$ |
|---|---|---|---|---|---|---|
| SDS [13] | 43.8 | 34.5 | 21.3 | 8.7 | 0.9 | - |
| Chen *et al*. [4] | 46.3 | 38.2 | 27.0 | 13.5 | 2.6 | - |
| PFN [24] | **58.7** | 51.3 | 42.5 | 31.2 | 15.7 | 52.3 |
| Ours | 58.3 | **52.4** | **45.4** | **34.9** | **20.1** | **53.1** |

Some success and failure cases of our method are shown in Figure 3. Our method has trouble differentiating instances which are occluded and also very visually similar to each other (Fig. 3, last two rows). The supplementary material includes a more detailed results table, and more examples of success and failure cases.

## 5 Conclusion

We have presented a simple method that effectively leverages state-of-the-art semantic segmentation and object detection networks to perform the less widely studied task of instance segmentation. Our approach begins with a bottom-up semantic segmentation of the input image, but is able to reason about instances since the final CRF layer in our semantic segmentation network incorporates information from an object detector in the form of a higher-order, detection potential. Our method produces state-of-the-art instance segmentation results, achieving considerable improvements over existing work at high $AP^r$ thresholds. Our final Instance CRF – which is dynamic in its number of labels – is fully differentiable. This means that our neural network can be dynamically instantiated for every input image and trained end-to-end.

#### Acknowledgements

We thank Bernardino Romera-Paredes, Stuart Golodetz and Shuai Zheng for insightful discussions and feedback. This work was supported by the EPSRC, Clarendon Fund, ERC grant ERC-2012-AdG 321162-HELIOS, EPRSRC grant Seebibyte EP/M013774/1 and EPSRC/MURI grant EP/N019474/1.

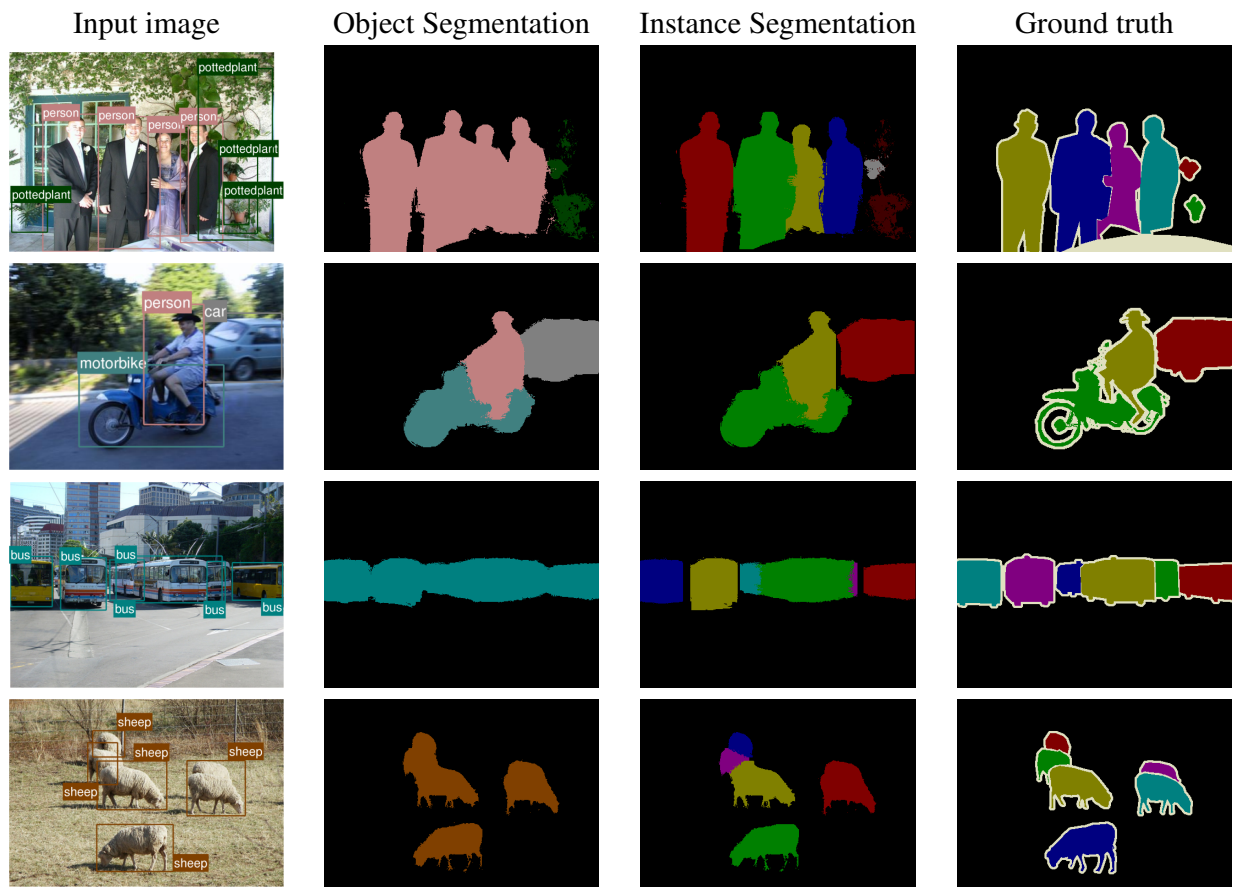

Figure 3: Some results of our system. Detector outputs are overlayed on the input image. Top row: An example where our method performs well, handling occlusions and false-positive detections. Second row: Our method performs well on this easy example which has no occluding objects of the same class. The naïve method described in §3.3 would have produced the same result. Third row: A failure case where the system is not able to differentiate visually similar instances. Bottom row: Heavily occluded objects are difficult to identify.

ARNAB AND TORR: BOTTOM-UP INSTANCE SEGMENTATION          11[7] Jifeng Dai, Kaiming He, and Jian Sun. Instance-aware semantic segmentation via multi-task network cascades. In *CVPR*, 2016.

[8] Mark Everingham, Luc Van Gool, Christopher KI Williams, John Winn, and Andrew Zisserman. The pascal visual object classes (voc) challenge. *IJCV*, 2010.

[9] Pedro Felzenszwalb, David McAllester, and Deva Ramanan. A discriminatively trained, multiscale, deformable part model. In *CVPR*, pages 1–8. IEEE, 2008.

[10] Spyros Gidaris and Nikos Komodakis. Object detection via a multi-region and semantic segmentation-aware cnn model. In *ICCV*, pages 1134–1142, 2015.

[11] Ross Girshick. Fast r-cnn. In *ICCV*, 2015.

[12] Ross Girshick, Jeff Donahue, Trevor Darrell, and Jitendra Malik. Rich feature hierarchies for accurate object detection and semantic segmentation. In *CVPR*, 2014.

[13] Bharath Hariharan, Pablo Arbeláez, Ross Girshick, and Jitendra Malik. Simultaneous detection and segmentation. In *ECCV*, pages 297–312. Springer, 2014.

[14] Bharath Hariharan, Pablo Arbeláez, Ross Girshick, and Jitendra Malik. Hypercolumns for object segmentation and fine-grained localization. *CVPR*, 2015.

[15] Kaiming He, Xiangyu Zhang, Shaoqing Ren, and Jian Sun. Spatial pyramid pooling in deep convolutional networks for visual recognition. In *ECCV*, 2014.

[16] Xuming He, Richard S Zemel, and Miguel Á Carreira-Perpiñán. Multiscale conditional random fields for image labeling. In *CVPR*, volume 2, pages II–695. IEEE, 2004.

[17] Pushmeet Kohli, Lubor Ladicky, and Philip H.S. Torr. Robust higher order potentials for enforcing label consistency. *IJCV*, 82(3):302–324, 2009.

[18] P. Krähenbühl and V. Koltun. Efficient inference in fully connected CRFs with Gaussian edge potentials. In *NIPS*, 2011.

[19] Alex Krizhevsky, Ilya Sutskever, and Geoffrey E. Hinton. Imagenet classification with deep convolutional neural networks. In *NIPS*, pages 1097–1105. 2012.

[20] Lubor Ladicky, Chris Russell, Pushmeet Kohli, and Philip HS Torr. Graph cut based inference with co-occurrence statistics. In *ECCV*, pages 239–253. 2010.

[21] L'ubor Ladický, Paul Sturgess, Karteek Alahari, Chris Russell, and Philip H. S. Torr. What, where and how many? combining object detectors and crfs. In *ECCV*, pages 424–437, 2010.

[22] Yann LeCun, Léon Bottou, Yoshua Bengio, and Patrick Haffner. Gradient-based learning applied to document recognition. *Proceedings of the IEEE*, 86(11):2278–2324, 1998.

[23] Xiaodan Liang, Yunchao Wei, Xiaohui Shen, Zequn Jie, Jiashi Feng, Liang Lin, and Shuicheng Yan. Reversible recursive instance-level object segmentation. *arXiv preprint arXiv:1511.04517*, 2015.